\def\year{2022}\relax
\documentclass[letterpaper]{article} 
\usepackage{aaai22}  
\usepackage{times}  
\usepackage{helvet}  
\usepackage{courier}  
\usepackage[hyphens]{url}  
\usepackage{graphicx} 
\def\UrlFont{\rm}  
\usepackage{natbib}  
\usepackage{caption} 
\DeclareCaptionStyle{ruled}{labelfont=normalfont,labelsep=colon,strut=off} 
\usepackage[utf8]{inputenc} 
\usepackage[T1]{fontenc}    
\usepackage{hyperref}       
\usepackage{url}            
\usepackage{tabularx}
\usepackage{wrapfig}
\usepackage{booktabs}       
\usepackage{amsfonts}       
\usepackage{nicefrac}       
\usepackage{microtype}      
\usepackage{xcolor}         
\usepackage{microtype}
\usepackage{comment}
\usepackage{graphicx}
\usepackage{booktabs} 
\usepackage{times}
\usepackage{epsfig}
\usepackage[utf8]{inputenc} 
\usepackage[T1]{fontenc}    
\usepackage{url}            
\usepackage{booktabs}       
\usepackage{nicefrac}       
\usepackage{microtype}      
\usepackage{microtype}
\usepackage[export]{adjustbox}
\usepackage{comment}
\usepackage{booktabs} 
\usepackage{amsmath}
\usepackage{subcaption}
\usepackage[vlined,ruled]{algorithm2e}
\usepackage{hyperref}       
\usepackage{amsfonts}       
\usepackage{nicefrac}       
\usepackage{microtype}      
\usepackage{verbatim}
\usepackage{times}
\usepackage{epsfig}
\usepackage{graphicx}
\usepackage{amsmath}
\usepackage{amssymb}
\usepackage{amsthm}
\usepackage{multirow}
\usepackage{hyperref}
\usepackage{multicol}
\usepackage{comment}
\usepackage{caption,subcaption}
\usepackage{footmisc}

\newtheorem{theorem}{Theorem}

\usepackage{newfloat}
\usepackage{listings}
\title{A Multi-agent Reinforcement Learning Approach for Efficient Client Selection in Federated Learning}

\author{
    Sai Qian Zhang\textsuperscript{\rm 1}, Jieyu Lin\textsuperscript{\rm 2}, Qi Zhang\textsuperscript{\rm 3}\\
}
\affiliations{
    \textsuperscript{\rm 1}Harvard University,
    \textsuperscript{\rm 2}University of Toronto,
    \textsuperscript{\rm 3}Microsoft

}

\usepackage{bibentry}

\begin{document}

\maketitle

\begin{abstract}
Federated learning (FL) is a training technique that enables client devices to jointly learn a shared model by aggregating locally-computed models without exposing their raw data. While most of the existing work focuses on improving the FL model accuracy, in this paper, we focus on the improving the training efficiency, which is often a hurdle for adopting FL in real-world applications. Specifically, we design an efficient FL framework which jointly optimizes model accuracy, processing latency and communication efficiency, all of which are primary design considerations for real implementation of FL. Inspired by the recent success of Multi-Agent Reinforcement Learning (MARL) in solving complex control problems, we present \textit{FedMarl}, an MARL-based FL framework which performs efficient run-time client selection. Experiments show that FedMarl can significantly improve model accuracy with much lower processing latency and communication cost.
\end{abstract}

\section{Introduction}

The rapid adoption of Internet of Things (IoT) in recent years has resulted in tremendous growth of data (e.g., text, image, audio) generated on client devices. 
With the success of deep neural networks (DNNs), there is a growing demand for efficiently training DNN models using the massive volume of data generated from client devices. 
The traditional centralized approach of gathering and training with all data at a central location not only raises scalability challenges, but also exposes data privacy concerns. To address these limitations, 
Federated Learning (FL)~\cite{mcmahan2017communication} has been recently proposed to allow client devices to jointly learn a shared model by aggregating their local DNN weight updates without exposing their raw data. At the beginning of each training round, a central server selects a subset of client devices to receive the current global model. Next, each client device performs DNN training using its local dataset and sends its weight update to the central server. The central server then applies the weight updates received from client devices to the global model, which will be used as the initial model for the next training round. 
Unlike the centralized training scheme, FL naturally addresses the scalability and privacy issues, and hence is more applicable for real implementation.

Despite its advantages, training DNNs using FL still faces several challenges. 
First, the underlying statistical 
heterogeneity leads to non-independent, identically distributed (non-IID) training data, which can severely degrade FL convergence behavior~\cite{zhao2018federated}.
Second, from the implementation perspective, the heterogeneous computing and networking resources on client devices can lead to the stragglers that will significantly slow down the FL training process.
This problem is further exacerbated by the uneven distribution of client data size, as a client with more data generally incurs higher training latency. Lastly, FL training process often causes high communication cost due to the iterative model updates between client devices and the central server.
From the practical perspective, an efficient FL framework that can mitigate all the three problems above is of paramount importance. While numerous solutions have been proposed for accelerating the FL convergence speed~\cite{karimireddy2019scaffold,li2020federated,wang2020optimizing,yang2021achieving}, minimizing the processing latency~\cite{wang2020optimize,nishio2019client,dhakal2019coded,wang2020tackling}, or reducing the communication overhead~\cite{konevcny2016federated,singh2019detailed,luping2019cmfl,sattler2019robust}, these heuristic solutions do not tackle these problems jointly. Worse still, optimizing one of the objectives might deteriorate the others. For instance, Scaffold~\cite{karimireddy2019scaffold} enables a better convergence behavior by computing additional model gradients, resulting in a large training latency.

To address this limitation, in this paper, we present a novel FL framework which jointly alleviates the three problems above, namely model accuracy, processing latency and communication overhead. 
Designing such an optimal client selection policy is challenging. This is due to the intractable convergence behavior of the FL training with the non-IID client data, the dynamic system performance on client devices and the complicated interaction between these objectives. More importantly, the significance of each optimization goal also varies across different scenarios. 
For example, communication cost may be the primary target in scenarios where network connection is costly, whereas minimizing total processing time may become the key objective for a delay-sensitive FL application.
Therefore, the proposed solution must balance and offset different design goals. Developing such a hand-tuned heuristic solution for each combination of these goals requires substantial time and effort. 

In this work, we leverage the recent advances in Multi-Agent Reinforcement Learning (MARL), and propose \textit{FedMarl}, an efficient FL framework that performs optimal client selection at run time. Although it is also possible to solve this problem with conventional reinforcement learning approach (i.e. single-agent RL), the high dimensionality on the action space for the RL agent will seriously degrade the convergence speed of RL training, leading to a suboptimal performance. FedMarl exploits the device performance statistics and training behaviors to produce efficient client selection decisions based on the designated objective imposed by the application designers. To closely simulate the real FL system operations, we perform extensive experiments to collect the real traces on DNN training latencies and transmission latencies over multiple mobile devices. We further build an MARL environment using the collected traces 
\begin{figure}
  \centering
  \includegraphics[width=0.35\textwidth]{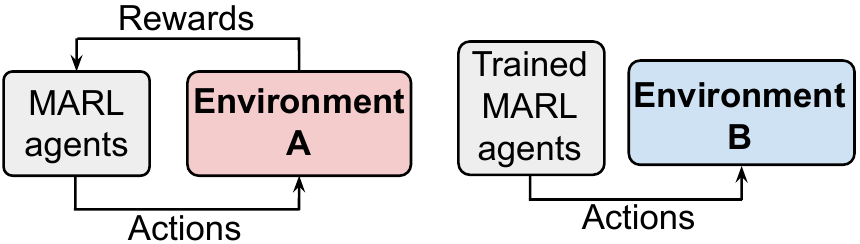}
  \caption{The MARL agents trained under one FL setting (left) can perform well under different FL settings (right).}
  \label{fig:FedMarl-dynamic}
\end{figure}
for the training of the MARL agents. Compared with the other algorithms, FedMarl achieves the best model accuracy while greatly reducing the processing latency and communication cost by $1.7\times$ and $2.2\times$, respectively. Furthermore, the evaluation results show that the superior performance of the MARL agents can translate to different FL settings without retraining (Figure~\ref{fig:FedMarl-dynamic}), which significantly broadens the applicability of the FedMarl. 

\section{Background and Related Work}

\subsection{Federated Learning}
\label{sec:fedavg}
The first FL framework, Federated Averaging (FedAvg)~\cite{mcmahan2017communication}, uses a central server to communicate with clients for decentralized training.
Specifically, given a total of $K$ client devices, during each training round $t$, a fixed number of $N$ $(N\leq K)$ devices are randomly picked by the central server. Each selected client $n$ $(1\leq n \leq N)$ contains local training data of size $D_{n}$. During the FL operation, each client device 
receives a copy of the global DNN model $W_{glb}^{t}$ from the central server, 
performs $E$ epochs of local training using local surrogate of the global objective function $F_{n}(.)$, and transmit the weight updates $\Delta W^{t}_{n}$ back to the central server. The central server then computes the weighted average $\frac{\sum_{n\in N} D_{n}\Delta W^{t}_{n}}{\sum_{n\in N}D_{n}}$ and applies the change to $W_{glb}^{t}$, generating the global weight $W_{glb}^{t+1}$ for the next training round. 
Although FedAvg achieves superior performance on homogeneous clients with IID data, its performance degrades when data distribution among client devices is non-IID~\cite{zhao2018federated,wang2020optimizing}. 
While multiple FL frameworks were proposed to mitigate the impact of statistical diversity by either controlling the divergence between the local model and global model~\cite{li2020federated, karimireddy2019scaffold}, or by training a personalized model for each client~\cite{smith2017federated,deng2020adaptive,zhang2020personalized}, none of these studies has considered system performance objectives in their design. 

\begin{figure}
  \centering
    \begin{subfigure}{.23\textwidth}
        \centering
        \includegraphics[width=\linewidth]{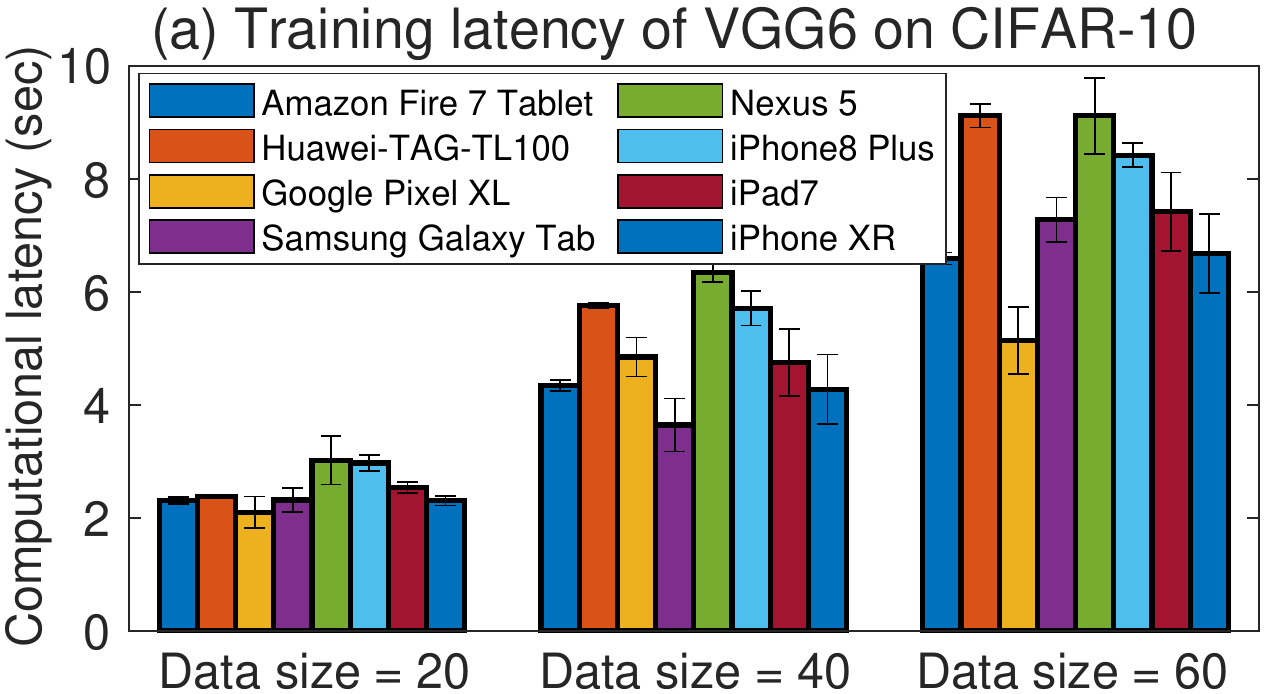}
    \end{subfigure}
    \begin{subfigure}{.23\textwidth}
        \centering
        \includegraphics[width=\linewidth]{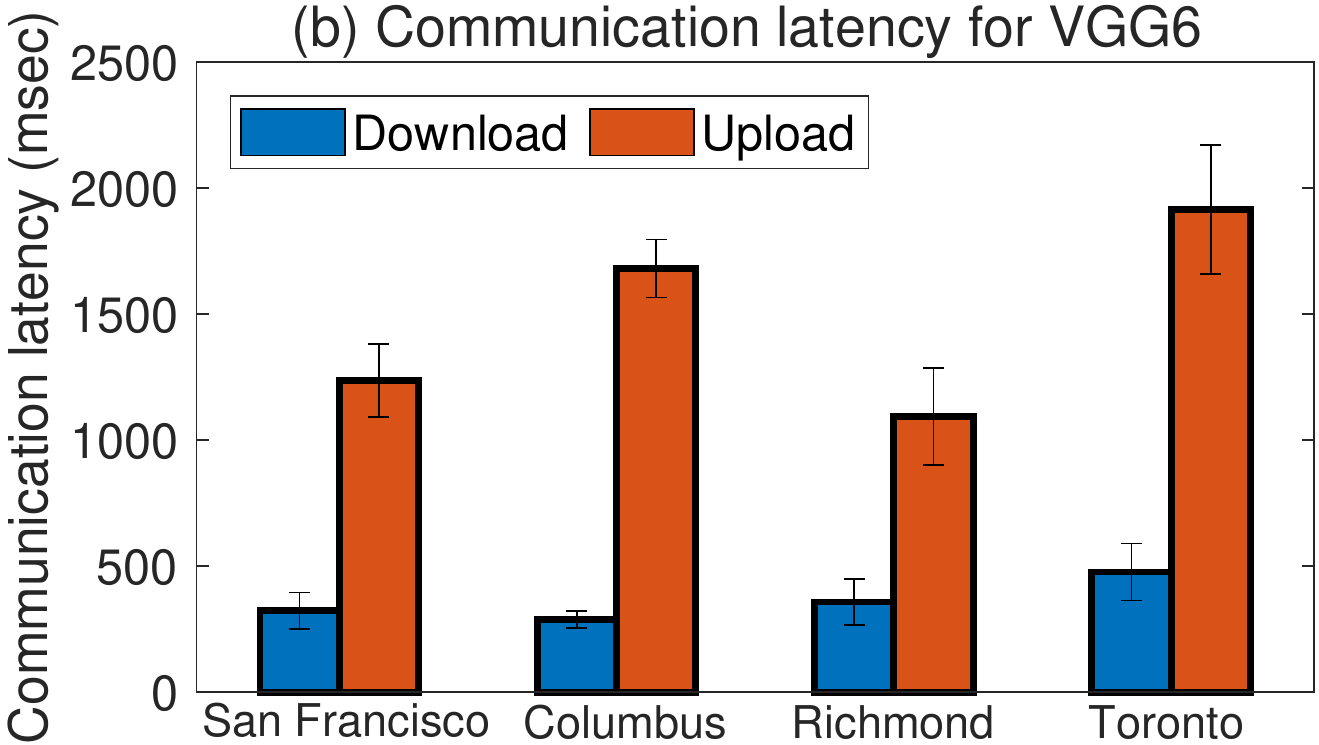}
    \end{subfigure}
    \caption{(a) Processing and (b) communication latencies for training VGG6 and CIFAR-10.}
    \label{fig:lenet-vgg6-computation-latency}
\end{figure}

To achieve optimal model performance with minimized processing latency, in \citet{wang2020optimize}, the authors proposed a FL framework to dynamically control the optimal partitioning of the client data.
In~\citet{nishio2019client}, the authors propose to mitigate the impact of stragglers by assigning higher priorities to faster clients devices. Besides latency reduction, multiple solutions have been proposed for improving FL communication efficiency. Most of these studies reduce the communication cost by applying filtering algorithms or quantization techniques to eliminate the unimportant weight updates~\cite{konevcny2016federated,luping2019cmfl,lai2020oort,fraboni2021clustered,sattler2019robust,reisizadeh2020fedpaq}, or compressing the weight updates via sketching ~\cite{rothchild2020fetchsgd}. While these studies have achieved significant performance improvement on their corresponding objectives, none of them can jointly optimize all the three objectives. 

\subsection{Multi-Agent Reinforcement Learning}
\label{sec:marl-description}
In cooperative MARL, a set of $N$ agents are trained to produce the optimal actions that lead to the maximum team reward. 
Specifically, at each timestamp $t$, each agent $n (1\leq n\leq N)$ observes its state $s^{t}_{n}$ and selects an action $a^{t}_{n}$ based on $s^{t}_{n}$. After all the agents complete their actions, the team receives a joint reward $r_{t}$ and proceeds to the next state $s^{t+1}_{n}$. The goal is to maximize the total expected discounted reward $R = \sum_{t=1}^{T} \gamma^{t} r_{t}$ by selecting the optimal agent actions, where $\gamma\in [0,1]$ is the discount factor. 
Recently, Value Decomposition Network (VDN)~\cite{sunehag2017value} has become a promising solution for jointly training agents in cooperative MARL. In VDN, each agent $n$ uses a DNN to infer its action. This DNN implements the Q-function $Q^{\theta}_{n}(s,a) = E[R_{t}|s^{t}_{n} = s, a^{t}_{n} = a]$, where $\theta$ is a parameter of the DNN and $R_{t}=\sum_{i=t}^{T} \gamma^{i}r_{i}$ is the total discounted team reward received at $t$. During MARL execution, every agent $n$ selects the action $a^{*}$ with the maximum Q-value (i.e., $a^{*} = \arg\max_{a} Q^{\theta}_{n}(s^{t}_{n},a)$). 
To train the VDN, a replay buffer is used to save the transition tuples $\big \langle s^{t}_{n},a^{t}_{n},s^{t+1}_{n},r_{t}\big \rangle$ for each agent $n$. A joint Q-function $Q_{tot}(.)$ is represented as the elementwise summation of all the individual Q-functions (i.e., $Q_{tot}(\textbf{s}_{t},\textbf{a}_{t}) = \sum_{n}Q^{\theta}_{n}(s_{n}^{t}, a_{n}^{t})$), where $\textbf{s}_{t} = \{s_{n}^{t}\}$ and $\textbf{a}_{t} = \{a_{n}^{t}\}$ are the states and actions collected from all agents $n\in N$ at timestep $t$. The agent DNNs can be trained recursively by minimizing the loss $L = E_{\textbf{s}_{t},\textbf{a}_{t},r_{t},\textbf{s}_{t+1}}[y_{t}-Q_{tot}(\textbf{s}_{t},\textbf{a}_{t})]^{2}$, where $y_{t} = r_{t} + \gamma \sum_{n} max_{a} Q^{\theta'}_{n}(s^{t+1}_{n},a)$ and $\theta'$ represents the parameters of the \emph{target network}, which are copied periodically from $\theta$ during the training phase. 
\begin{figure}
  \centering
    \begin{subfigure}{.23\textwidth}
        \centering
        \includegraphics[width=\linewidth]{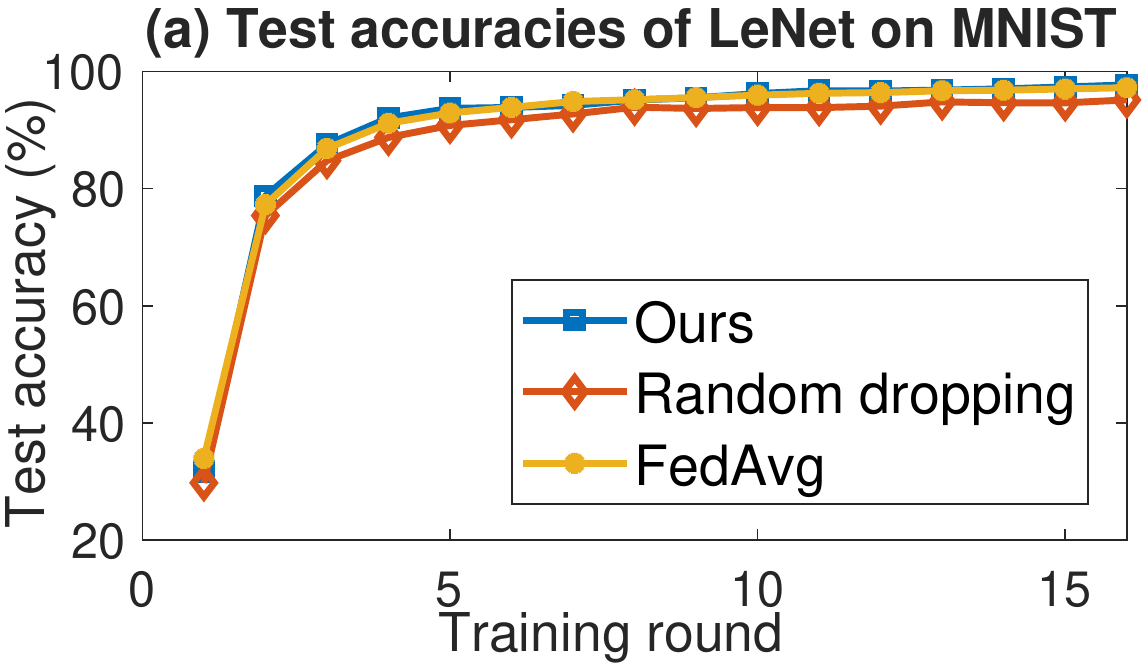}
    \end{subfigure}
    \begin{subfigure}{.23\textwidth}
        \centering
        \includegraphics[width=\linewidth]{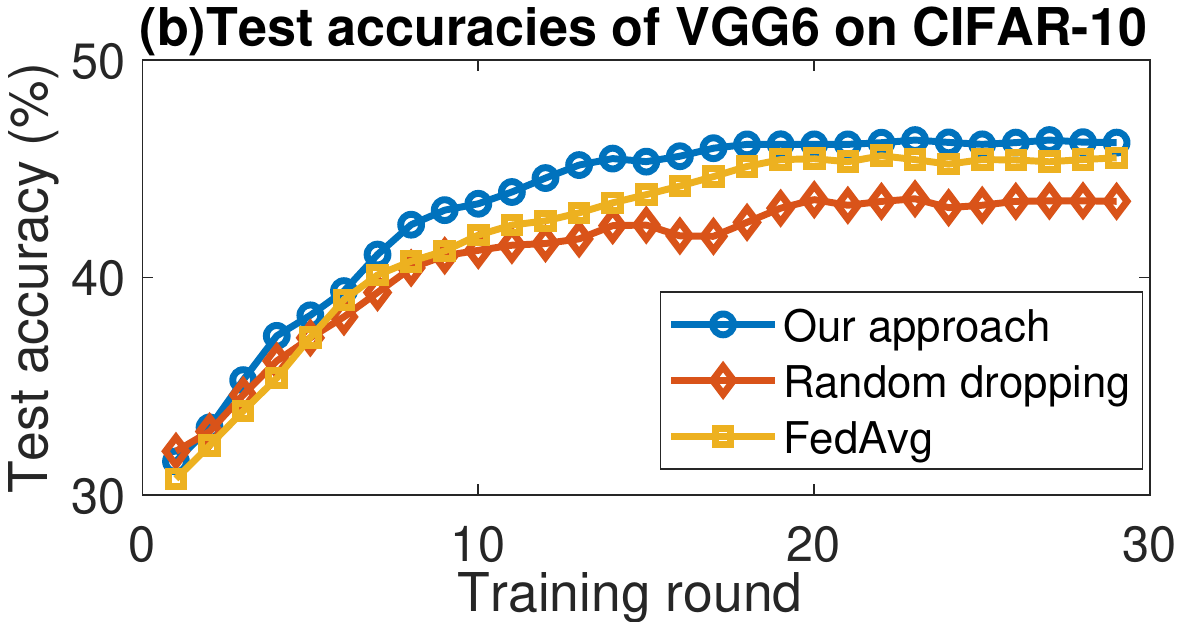}
    \end{subfigure}
    \caption{Test accuracies of the three algorithms on (a) LeNet and (b) VGG6.}
    \label{fig:acc-motivation-performance}
\end{figure}

\section{Motivation and Problem Definition}
\subsection{FL Processing Time Breakdown}
\label{sec:latency-measurement}
To better understand the composition of the FL processing time, we trace the processing time for training VGG6~\cite{simonyan2014very} by using CIFAR-10~\cite{cifar-dataset} dataset. To simulate resource heterogeneity, we use $8$ different client devices, including iPhone8, iPhone XR, iPad2, Huawei-TAG-TL100, Google Pixel XL, Nexus 5, Samsung Galaxy Tab A 8.0 and Amazon Fire 7 Tablet. For each device, we measure the time for performing $5$ epochs of training with different data size ranging from $20$ to $60$ samples. We developed our own DNN training implementation on the mobile devices using the previous literature~\citep{TFtraining,CoreMLtraining}. For the Android devices, the DNN models are first built with Keras and then converted to TensorFlow Lite format~\cite{tflite} before running on the client device. For iOS devices, we first build the DNN models using PyTorch and convert the resulting model to CoreML~\cite{core-ml} format with coremltools~\cite{coreml-tool}. We then perform the measurement 500 times and record the average latency. The measurement are shown in Figure~\ref{fig:lenet-vgg6-computation-latency} (a). 
To measure the communication latency between the central server and client devices, we use a single-core Amazon EC2 \textit{p3.2xlarge} instance located at Portland (OR) to simulate the cloud server, and four virtual machines (VMs) located at Richmond (VA), San Francisco (CA), Columbus (OH), and Toronto (ON) to simulate the client devices. 
Figure~\ref{fig:lenet-vgg6-computation-latency} (b) shows the average latency for uploading and downloading VGG6 models over 500 measurements. 

We make the following observations from Figure~\ref{fig:lenet-vgg6-computation-latency}. First, the majority of processing time is spent on DNN training on the client devices. For example, the communication latency for uploading VGG6 is only $1.4s$ for the client device in Richmond, 
whereas training VGG6 for five epochs on Nexus 5 takes $8.9s$ with 60 training samples, which is $6.4\times$ higher than the communication latency. Second, the training time varies significantly across client devices. For instance, training VGG6 with 60 samples for five epochs on Google Pixel XL takes only $4.8s$, which is $2\times$ faster than on Nexus 5. Finally, uploading DNN models from client devices is much slower than downloading DNN models from the central server. For example, downloading VGG6 to the client device in Columbus only takes $361ms$, whereas uploading the same DNN takes $1.68s$ ($4.7\times$ larger). Measurements on LeNet and ResNet-18 show a similar trend. 

\subsection{FL Convergence under Non-IID Client Data}
\label{sec:noniid-accuracy}
In this section, we study the influence of Non-IID data on the FL convergence and possible solutions to mitigate its impact. The unstable FL convergence behavior is triggered by the inconsistency among the local objective functions $F_{n}(.)$, which is caused by the non-IID-ness in the client training data. One promising approach to alleviate the disparity among local objective functions is to eliminate biased model updates, which are outliers that can hurt overall convergence rate. 
Previous literature has shown that excluding these outliers can significantly accelerate the training process~\cite{luping2019cmfl,duan2021feddna,abay2020mitigating}. For example, CMFL~\cite{luping2019cmfl} utilizes the total number of sign differences between the local and global model parameters to measure the bias of the local model. However, this requires all the clients to first finish their local training to produce the local DNN weights, resulting in high processing latency. Moreover, counting the total sign differences will introduce additional computation overhead and further deteriorates the processing latency. 
To mitigate this issue, we measure the degree of bias using initial training loss, which is the training loss after the first epoch of local training process at each client. Using initial training loss as the indicator offers several advantages. First, the initial training loss naturally reflects the degree of inconsistency between the local client data and the global model, which can be further used to estimate the degree of bias on the local updates. 
Second, the initial training loss is produced as an intermediate result without any additional computational overhead. After the first epoch, each client device reports its initial training loss to the central server. The server collects the losses and performs early rejection by halting the remaining training process on the devices with high losses. Sending the initial training losses to central server will not impair the processing latency and communication cost, as the training loss (a single scalar) is tiny compared to the DNN model. Conversely, this approach will lower the processing latency and communication overhead, because only a subset of the clients need to complete their local training processes and send their weight updates. For simplicity, we call the first epoch of the local training process the \textit{probing training}, and the initial training loss the ~\textit{probing loss}. 
\begin{figure}
  \centering
  \includegraphics[width=0.4\textwidth]{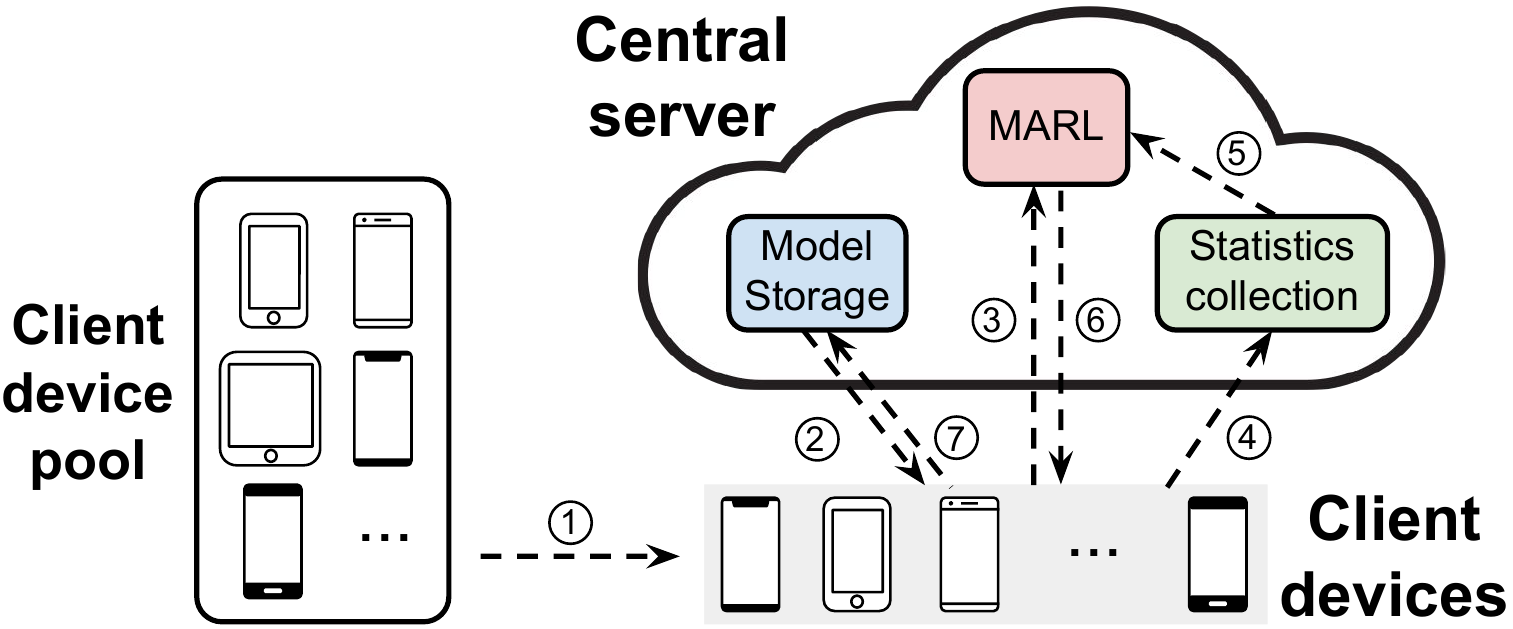}
  \caption{FedMarl workflow, steps are shown in circles.}
  \label{fig:FedMarl-system}
\end{figure}
We compare our bias estimation approach with two baseline approaches. The first approach eliminates the outliers by randomly dropping $50\%$ of the clients. The second approach implements FedAvg by keeping all the devices in the local training process. For our approach, at each training round, we reject half of the clients with their probing loss higher than the average probing loss. Figure~\ref{fig:acc-motivation-performance} depicts the average test accuracies on LeNet and VGG6. Our method achieves a higher convergence speed. Specifically, our method obtains test accuracies of $98.7\%$ and $45.2\%$ on MNIST and CIFAR-10, which are $1.3\%$ and $2.8\%$ higher than the other methods on average. The above results demonstrate that the probing loss can be used as \textbf{an indicator of client selection} for better FL accuracy under non-IID setting.

\subsection{Problem Formulation}

We present the FL optimization problem formulation in this section.
Let $H^{c}_{t,n}$ denote the local training time for client $n\in N$ at round $t\in T$. Also denote by $H^{u}_{t,n}$ the communication latency for uploading the local model from client $n$ to the central server. During each training round $t$, a subset of clients will be early rejected by the central server, while the rest clients will continue to finish their local training process. 
Define $B^{t}_{n}$ as the communication cost for sending updates from client $n$ to the central server at round $t$. Let $a^{t}_{n}\in \{0,1\}$ denote whether the client $n$ is chosen to complete full local training. The total processing latency $H_{t}$ of a training round $t$ and the total communication cost $B_{t}$ for round $t$ can be expressed as:
\begin{equation}
H_{t} = \max\limits_{1\leq n\leq N} (H^{u}_{t,n} + H^{c}_{t,n}) a^{t}_{n}
   \enspace\mathrm{and}\enspace 
B_{t} = \sum_{n} B^{t}_{n}a^{t}_{n}
\label{eqn:latency}
\end{equation}
Finally, let $Acc(T)$ denote the test accuracy of the global model on a global test dataset after the last training round $T$. Our FL system optimization problem can be defined as:
\begin{equation}
    \label{eqn:obj}
    \max\limits_{A} E\Big[w_{1}Acc(T) - w_{2}\sum_{t\in T} H_{t} - w_{3}\sum_{t\in T}  B_{t} \Big]
\end{equation}
where $A = [a^{t}_{n}]$ is a $T\times N$ matrix for client selection, $w_{1}$,$w_{2}$,$w_{3}$ are the importance of the objectives controlled by the FL application designers. Our objective is to maximize the accuracy of the global model while minimizing the total processing latency and communication cost. The expectation is taken over the stochasticity of DNN training.

\section{FedMarl System Design}
The FL optimization problem is difficult to solve directly. We instead model the problem as a MARL problem. In this section, we present our problem formulation and FedMarl system design.

\subsection{MARL Agents Training Process}
In FedMarl, each client device $n$ relies on an MARL agent at the central server to make its participation decision. Each MARL agent contains a simple two-layer Multi-layer perceptron (MLP) that is cheap to implement. During the training phase of MARL, each MARL agent takes its current state and infers its action $a^{t}_{n}$. Based on the client selection pattern, the central server computes the team reward by considering the test accuracy improvement on the global model, total processing latency and the communication cost. The MARL agents are then trained with VDN to maximize the team reward.

\subsubsection{Design of MARL Agent States}
\begin{table}[tp!]
\centering
\scriptsize
\resizebox{0.38\textwidth}{!}{
\begin{tabular}{| c | c |}
\hline
\multirow{ 3}{*}{Device type} & iPhone8, iPhoneXR, iPad2, \\
 &Huawei-TAG-TL100, Google Pixel XL, Nexus5, \\
 &Samsung Galaxy Tab A, Amazon Fire 7 Tablet \\ \hline
Data size & 20, 25, 30, 35, 40, 45, 50, 55, 60\\\hline
\multirow{ 2}{*}{Location}  & Richmond (Virginia), San Francisco (California), \\
 &Columbus (Ohio), and Toronto (Ontario)\\\hline
\end{tabular}}
\caption{Possible options for client device settings.}
\label{table:client-option}
\end{table}
The state of each MARL agent $n$ consists of six components: the probing loss, the processing latency of probing training, historical values on communication latency, the communication cost from a client device to server, the size of the local training dataset and the current training round index. Let $L^{t}_{n}$ denote the probing loss of agent $n$ in round $t$. At each round $t$, each agent first performs the probing training and sends $L^{t}_{n}$ to the central server. 
$L^{t}_{n}$ is then examined by the corresponding MARL agent in the central server to infer its degree of bias on the local model updates. In addition, to infer the current training latency and communication latency at client device $n$, each MARL agent is provided with the historical probing training latencies $\textbf{H}_{t,n}^{p} = [H^{p}_{t-\Delta T_{p},n},...,H^{p}_{t,n}]$ and communication latencies $\textbf{H}_{t,n}^{u} = [H^{u}_{t-\Delta T_{c},n},...,H^{u}_{t-1,n}]$, where $H^{p}_{t,n}$ and $H_{t,n}^{u}$ denote the latencies for probing training and model uploading of agent $n$ at training round $t$. $\Delta T_{p}$ and $\Delta T_{c}$ are the sizes of the historical information. Finally, each MARL agent $n$ also involves the communication cost $B^{t}_{n}$, the local data size $D_{n}$ and training round index $t$ in its input state. This is because the individual communication cost contributes to the total communication cost, and the training data size affects training latency and model accuracy. The state vector $\textbf{s}^{t}_{n}$ of agent $n$ at round $t$ is defined as:
\begin{equation}
\label{eqn:state}
    \textbf{s}^{t}_{n} = [L^{t}_{n}, \textbf{H}_{t,n}^{p}, \textbf{H}_{t,n}^{u}, B^{t}_{n}, D_{n}, t]
\end{equation}
To accelerate the training of VDN, we normalize each element in the state vector to make them under the same scale. In addition, this will improve the generalizability of the MARL agents under different system conditions, as shown in the evaluation section. Finally, all the MLPs in the MARL agents share their weights in order to reduce the storage overhead and prevent lazy agent problem~\cite{jiang2018learning}.  

\subsubsection{Description on Agent Actions}
Given the input state shown in equation~\ref{eqn:state}, each MARL agent $n$ decides whether the client device $n$ should be terminated earlier. In particular, the MARL agent produces a binary action $a^{t}_{n}\in [0,1]$, where $a^{t}_{n}=0$ indicates the client device will be terminated after the probing training, and vice versa.
\begin{table}
\centering
\resizebox{0.45\textwidth}{!}{%
    
    \begin{tabular}{c | c | c | c | c }
    \hline
            & MNIST & CIFAR-10 & F-MNIST & Shakespeare  \\ \hline
    \textbf{FedMarl} & \textbf{96.91\%}  &  \textbf{48.87\%}  &  \textbf{96.14\%}  &  \textbf{44.58\%}  \\ 
    FedAvg & 94.40\%  &  43.49\%  &  94.70\%  &  40.66\% \\
    HeteroFL & 96.86\%  &  47.74\%  &  96.08\%  &  43.98\% \\
    FedProx & 95.85\%  &  47.32\%  &  95.73\%  &  43.66\% \\ 
    FedProx-THS & 94.43\%  &  45.51\%  &  94.80\%  &  42.69\% \\
    FedNova & 96.07\%  &  47.97\%  &  95.93\%  &  44.10\% \\ 
    FedNova-THS & 95.30\%  &  44.88\%  &  94.42\%  &  42.41\% \\
    Oort & 96.09\%  &  48.11\%  &  96.02\%  &  44.05\% \\ 
    CS & 95.89\%  &  48.19\%  &  96.14\%  &  43.97\% \\ \hline
    \end{tabular}
    }
    \caption{Accuracy performance of LeNet, VGG6, ResNet-18, LSTM on their datasets. FedMarl achieves the best accuracies across all the datasets.}
\label{tal:acc-performance}
\end{table}

\subsubsection{Design of the Reward Function}
To optimize the FL performance described in Equation~\ref{eqn:obj}, the reward function should reflect the changes in the test accuracy, processing latency and communication cost after executing client selection decisions generated by the MARL agents. The reward $r_{t}$ at training round $t$ is defined as:
\begin{equation}
    \label{eqn:marl-reward}
    r_{t} = w_{1}\Big[U(Acc(t)) - U(Acc(t-1))\Big] - w_{2}H_{t} - w_{3}B_{t}
\end{equation}
$H_{t}$ is the processing latency of round $t$ and is defined as:
\begin{equation}
    H_{t} = \max\limits_{1\leq n\leq N} (H^{p}_{t,n}) + \max\limits_{n:1\leq n\leq N, a^{t}_{n}=1} (H^{rest}_{t,n} + H^{u}_{t,n})
\end{equation}

Here, $\max_{1\leq n\leq N} H^{p}_{t,n}$ represent the total time needed for generating all the probing losses. The MARL agents utilize these probing losses to select the client devices that will continue the local training and upload their model updates, which needs an additional time of $\max_{n:1\leq n\leq N, a^{t}_{n}=1} (H^{rest}_{t,n} + H^{u}_{t,n})$, where $H^{rest}_{t,n}$ is the time required for client device $n$ to finish the local training process.
Moreover, $U(.)$ is a utility function that ensures $U(Acc(t))$ can still alter moderately even if $Acc(t)$ improvement is small near the end of the FL process. $B_{t}$ is the total communication cost as defined in equation~\ref{eqn:latency}. The MARL agents are trained using VDN described in background section.

\subsection{FedMarl System Workflow}

Figure~\ref{fig:FedMarl-system} provides an overview of the FedMarl workflow. The central server contains three building blocks: \textit{Model storage block} for storing and updating the global DNN model, \textit{MARL block} for executing the trained MARL agents and generating the client selection decision, and the \textit{Statistics collection block} for gathering the client device statistics such as processing latency and communication latency. At each training round, $N$ client devices are picked from the client device pool with the criteria described in~\cite{bonawitz2019towards} (step 1). The selected devices then receive a copy of global DNN model from the Model storage block (step 2). Next, the client devices perform the probing training and send their probing losses to the MARL block (step 3). Meanwhile, the client devices also transmit their probing training latencies to the Statistics collection block (step 4). After receiving all the probing losses from the clients, the MARL agents take these losses together with the historical information from the Statistics collection block (step 5) and produce the client selection decisions $a^{t}_{n}$ (step 6). The selected client devices then perform the rest local training and deliver their model updates to the Model storage block (step 7), which will apply the weight updates to the global DNN model. Meanwhile, the Statistics collection block also updates the existing statistics on communication latency. 

\begin{figure}[tp!]
    \centering
    \begin{subfigure}{.23\textwidth}
        \centering
        \includegraphics[width=0.95\linewidth]{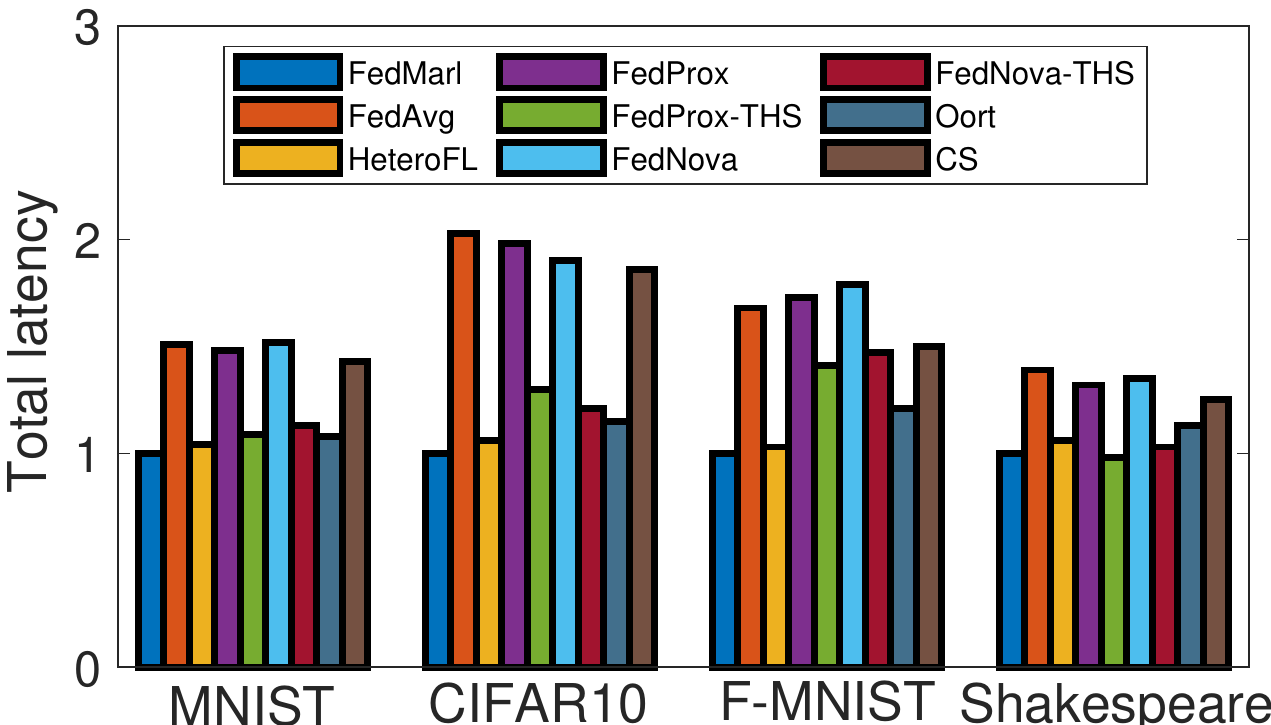}
    \end{subfigure}%
    \begin{subfigure}{0.23\textwidth}
        \centering
        \includegraphics[width=0.95\linewidth]{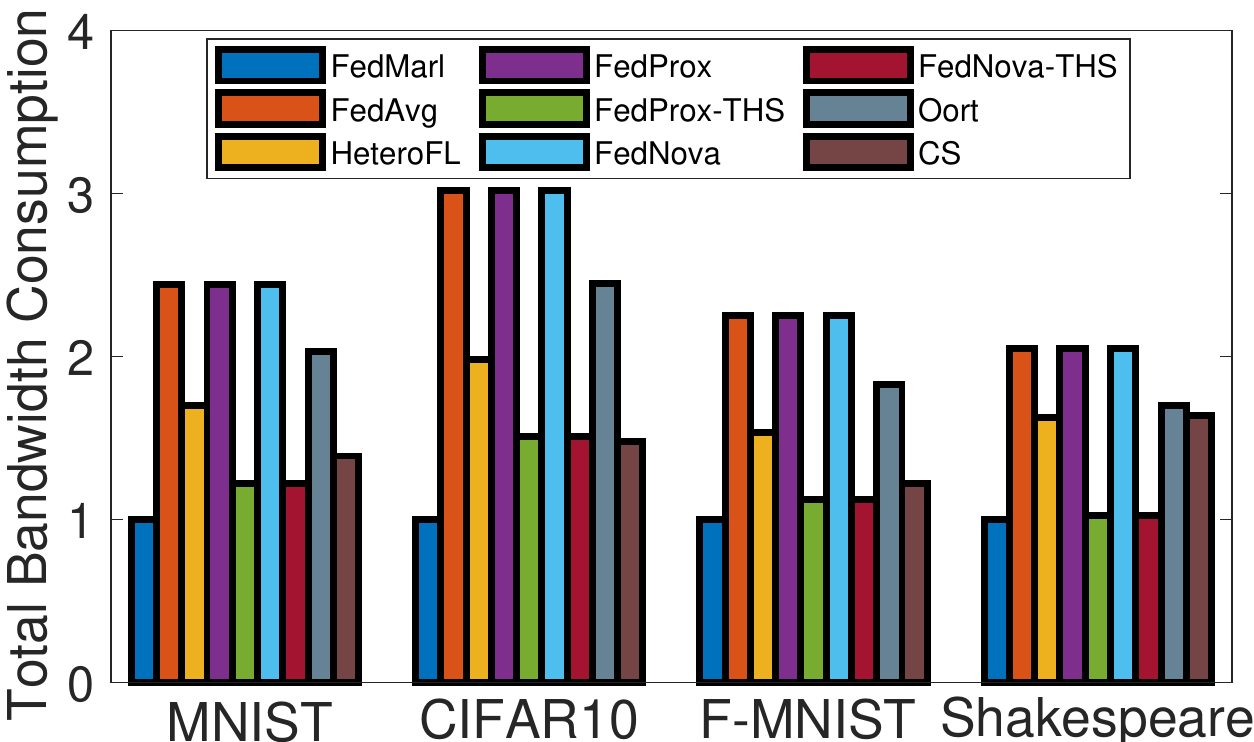}
    \end{subfigure}
    \caption{Comparison on (a) normalized total latency and (b) normalized communication cost.}
    \label{fig:latency-bandwidth}
\end{figure}

\section{Theoretical Analysis}
In this section, we present a theoretical analysis to show the stability of FedMarl. The analysis is based on the work in~\citet{zhao2018federated}. 
Assume each client $n\in N$ contains $D_{n}$ non-IID local training data with underlying probability $P_{n}(.)$. Denote $\textbf{x}$ and $y_{\textbf{x}}$ as the training data and its ground truth label. Also denote $f_{m}(\textbf{x},W)$ as the output probability for input $\textbf{x}$ to take label $m$ by using DNN model $f$ . We further build a dataset by gathering all the Non-IID training data from every client, and assume the aggregated dataset to be IID approximately. Let $P(m)$ denote the probability of occurrence of label $m$ in this aggregated IID dataset. Let $V_{n}^{t,e}$ denote the local model weight of client $n$ at local epoch $e$ of training round $t$, and $V_{iid}^{j}$ the trained DNN weights at epoch $j$ by conventional SGD using the aggregated IID dataset. Theorem~\ref{thm:convergence} offers a bound on $||W_{glb}^{t}-V_{iid}^{Et}||$, where $W_{glb}^{t}$ is the global DNN model at central server after training round $t$, $V_{iid}^{Et}$ is the trained DNN weights at epoch $Et$ by using the aggregated dataset with conventional SGD.
\begin{theorem}
    Assume $\nabla_{W} E_{\textbf{x}|y_{\textbf{x}}=m}\log(f_{m}(\textbf{x},W))$ is $\sigma_{m}$-Lipschitz for each $m\in M$, assume $a^{t}_{n}$ satisfies $P(m) =  \sum_{n=1}^{N}\frac{D_{n}a^{t}_{n}P_{n}(m)}{\sum_{n=1}^{N}D_{n}a^{t}_{n}} + \epsilon_{t}$ for a constant $\epsilon_{t}$ and $\epsilon_{t}\leq \epsilon$ $\forall t$. We also assume that $||\nabla_{W} E_{\textbf{x}|y_{\textbf{x}}=m}\log(f_{m}(\textbf{x},W))|| \leq c_{max}$ $\forall \textbf{x}, m, W$. 
    Then
    $||W_{glb}^{t}-V_{iid}^{Et}|| \leq \sum_{n=1}^{N}q^{t}_{n} \Big[ (u_{n})^{E}||W^{t-1}_{glb}-V^{E(t-1)}_{iid}|| + \eta\epsilon NM c_{max} + \eta c_{max}\sum_{m=1}^{M}||P_{n}(m)-P(m)|| (\sum_{j=1}^{E-1}(u_{n})^{j})\Big] $,
    where $u_{n} = 1+\eta\sum_{m=1}^{M}P_{n}(m)\sigma_{m}$, $q^{t}_{n} = \frac{D_{n}a^{t}_{n}}{\sum_{n=1}^{N}D_{n}a^{t}_{n}}$ and $\eta$ is the learning rate.
    \label{thm:convergence}
\end{theorem}
Given the convergence of SGD, $V_{iid}^{Et}$ is bounded $\forall t\in T$. From the result of Theorem~\ref{thm:convergence}, we can see that $W_{glb}^{t}$ (model weights of FedMarl) is also bounded. Therefore FedMarl will converge. Proof is given in the supplementary materials.
\section{Evaluation}
\label{sec:evaluation-setting}

We evaluate FedMarl using several popular DNN models, including LeNet~\cite{lecun1998gradient} on MNIST~\cite{mnist-dataset}, VGG6~\cite{simonyan2014very} on  CIFAR-10~\cite{cifar-dataset} and ResNet-18~\cite{he2016deep} on Fashion MNIST~\cite{fashionmnist-dataset} for image classification, LSTM~\cite{hochreiter1997long} on Shakespeare 
dataset~\cite{shakespeare-dataset} for text prediction. To simulate device heterogeneity, each client device is randomly assigned a device type, a location and a training set size, as summarized in
Table~\ref{table:client-option}. We collect training latency data over the above DNNs for each device type under each data size, and measure the time for sending each DNN model from the client devices at each location to the central server, as described in the motivation section. 
To simulate the non-IID training data at each client device, we sort the training data by its label. For each client device, $80\%$ of its training data are from one random label, the rest of the training data are sampled uniformly from the remaining labels. The number of training data per device is generated by the power law~\cite{li2020federated}. All client devices adopt a uniform communication cost $B^{t}_{n} = 1$ $\forall t,n$. During each training round, $N=10$ client devices are randomly selected from a pool of $K=100$ devices. The training latency and communication latency for each device are sampled from the collected traces based on its device type, data size and location. Each client device first performs the probing training and reports its probing loss to the central server. Based on the feedback from the MARL agents, a subset of them will continue their local training process for $E=5$ epochs. The number of training round $T$ is set to $T=20$ for VGG6, and $T=15$ for LeNet, ResNet-18 and LSTM, respectively.
Each MARL agent consists of a MLP of two layers with a hidden layer of 256 neurons. For the reward function (equation~\ref{eqn:marl-reward}), we make $w_{1} = 1.0$,  $w_{2} = 0.2$ and $w_{3} = 0.1$. The utility function is defined to be $U(x) = 10-\frac{20}{1+e^{0.35(1-x)}}$ for shaping the test accuracy. The sizes of the historical information $\Delta T_{p}$ and $\Delta T_{c}$ are set to $3$ and $5$, respectively. We train the VDN with 300, 200, 300, 200 episodes for LeNet, VGG6, ResNet-18 and LSTM until convergence. To demonstrate the advantage of the MARL over the single-agent RL, we also implement the FedMarl with single-agent RL and compare their convergence behaviors. We observe that MARL approach converges at a much fast speed than single-RL approach under the same training environment. 

\begin{table}[tp!]
\centering
\scriptsize
\resizebox{0.45\textwidth}{!}{
\begin{tabular}{| c | c | c | c |}
\hline
{Source/Target} & LeNet & VGG6 & ResNet-18 \\ \hline
LeNet & 1.0 (1.0) & 0.720 (0.798) &  0.896 (0.930)    \\\hline
VGG6 & 0.906 (0.974) & 0.803 (0.803) & 0.859 (0.927)      \\\hline
ResNet-18 & 0.965 (0.989) & 0.731 (0.792) & 0.933 (0.933)    \\\hline
\end{tabular}
}
\caption{Performance of the trained MARL agents across different DNNs. All the results are normalized by the performance of agents trained with LeNet and evaluated on itself. Numbers in brackets show the performance after finetuning.}
\label{table:cross-dnn-evaluation}
\end{table}
\subsection{FedMarl Performance}
\label{sec:fedmarl-performance}
We compare the performance of FedMarl with multiple advanced benchmark algorithms, including: FedNova~\cite{wang2020tackling},  HeteroFL~\cite{diao2020heterofl}, Oort~\cite{lai2020oort}, FedProx~\cite{li2020federated}, Clustered Sampling (CS)~\cite{fraboni2021clustered} and FedAvg~\cite{mcmahan2017communication}. The clients perform local training for $E=6$ epochs in each training round. For FedProx, we adopt the optimal proximal term $\mu$ for each DNN, which gives $\mu= 1, 1, 0.1, 0.001$ for LeNet, VGG6, ResNet-18 and LSTM, respectively. For FedNova, the fast client will perform more local training steps until the slowest client finishes its training. The weight updates are then normalized based on the total number of local training steps. For HeteroFL, we adopt five computing complexity levels with a hidden channel shrinkage ratio of 0.5. For Oort, we 
set the exploitation factor step window and straggler penalty to 0.1,5,2 for all the tasks. For CS, we group the clients devices in the pool into 10 clusters, and one representative device is selected from a single cluster per training round.
In order to reduce the total processing latency, we further consider a simple client selection. Instead of performing the local training process on all the $N$ client devices, the new selection algorithm, called \textit{Top-half Speed} (THS), selects the bottom $50\%$ of the client devices with the lowest probing training latency during each training round. By removing the straggler client devices, THS can greatly lower the total processing latency and communication cost. We further apply THS on FedProx and FedNova, producing two additional benchmark algorithms: FedProx-THS and FedNova-THS. We perform the evaluation for 100 times and record the average performance for each algorithm. All the local training are performed with SGD.

Table~\ref{tal:acc-performance} and Figure~\ref{fig:latency-bandwidth} show the final test accuracy, total processing latencies and communication costs for all the algorithms. In Figure~\ref{fig:latency-bandwidth}, all values are normalized by the values of the FedMarl. FedAvg, FedProx and FedNova make all the clients transmit their local updates, therefore there is no reduction on processing latency and communication cost. In contrast, Oort, CS, HeteroFL, FedProx-THS and FedNova-THS enables only partial client to report their local updates, which reduces the communication cost and processing latency. We notice that FedMarl, HeteroFL and FedNova outperform the rest algorithms on test accuracy,
but FedMarl achieves the optimal accuracy, processing latency and communication cost at the same time. 

\begin{figure}
    \centering
    \begin{subfigure}{.23\textwidth}
        \centering
        \includegraphics[width=0.95\linewidth]{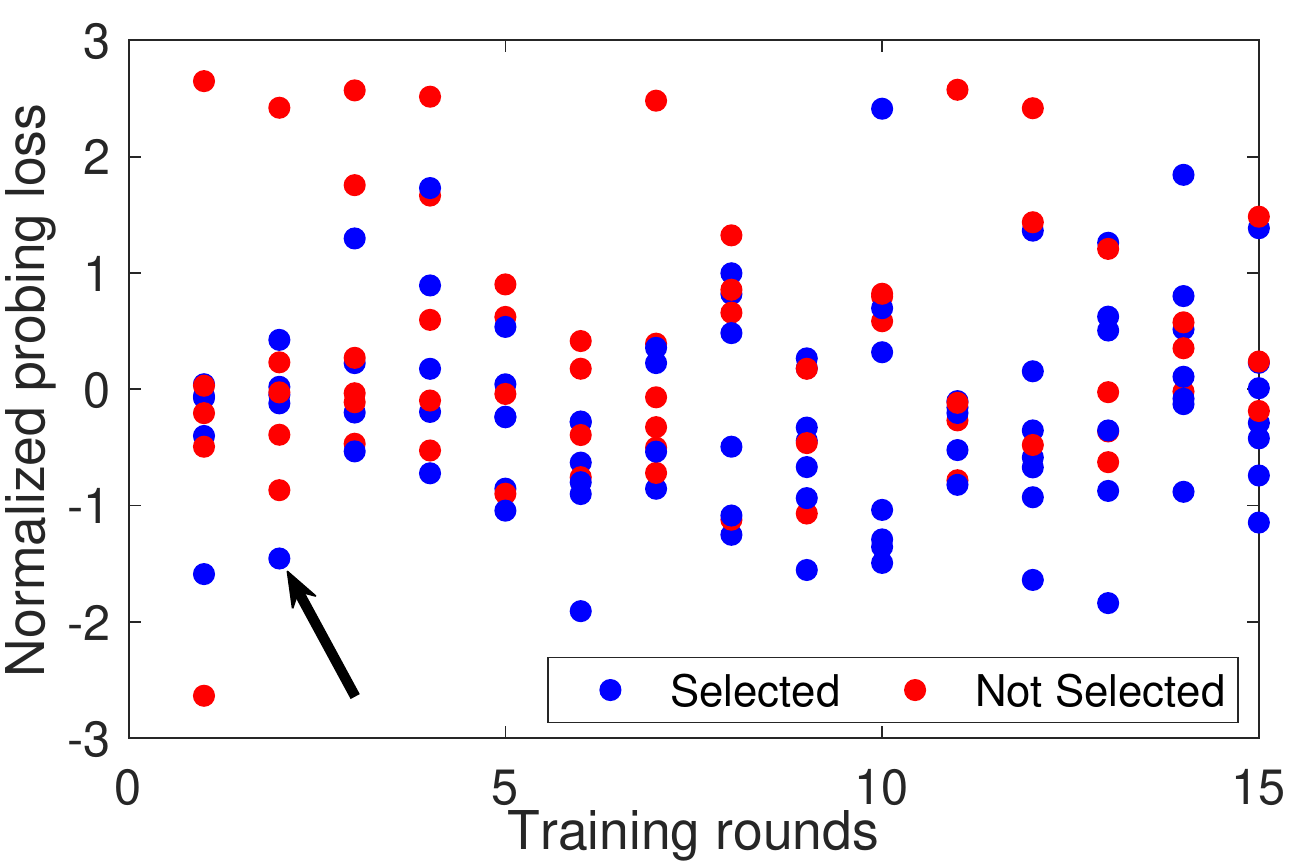}
    \end{subfigure}%
    \begin{subfigure}{0.23\textwidth}
        \centering
        \includegraphics[width=0.94\linewidth]{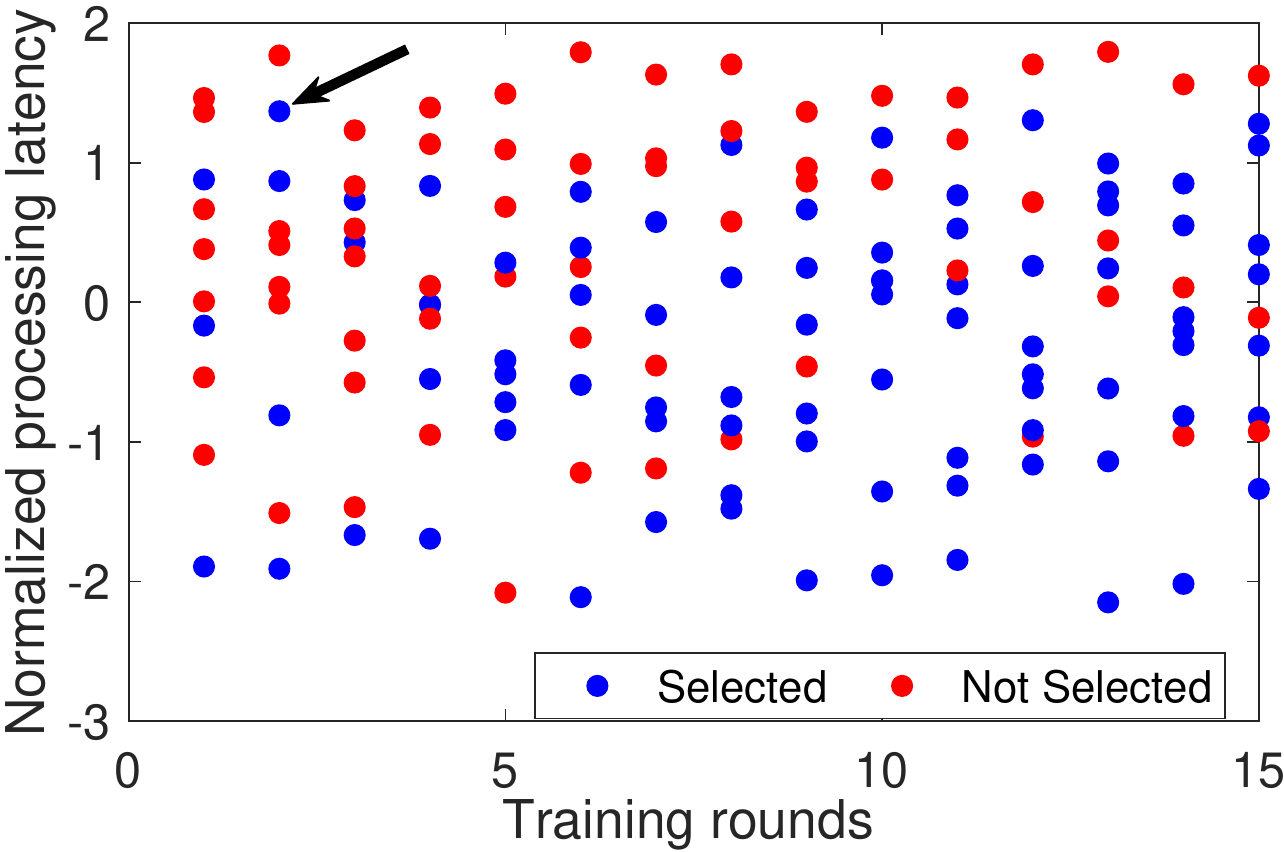}
    \end{subfigure}
    \caption{Decisions made by FedMarl. Each dot represent a client device. The blue dot means the client is selected for local training, the red dot means the client is early rejected.}
    \label{fig:scheduling}
\end{figure}

\begin{figure*}[tp!]
    \centering
    \begin{minipage}{0.69\textwidth}
    \begin{subfigure}{.33\textwidth}
        \centering
        \includegraphics[width=\linewidth]{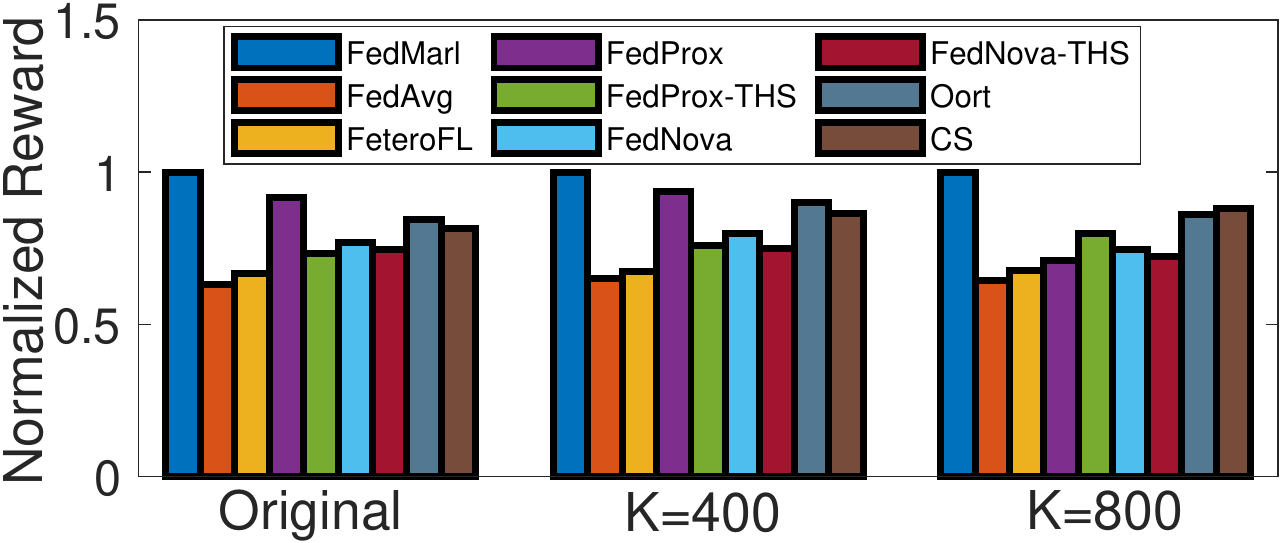}
    \end{subfigure}%
    \begin{subfigure}{.33\textwidth}
        \centering
        \includegraphics[width=\linewidth]{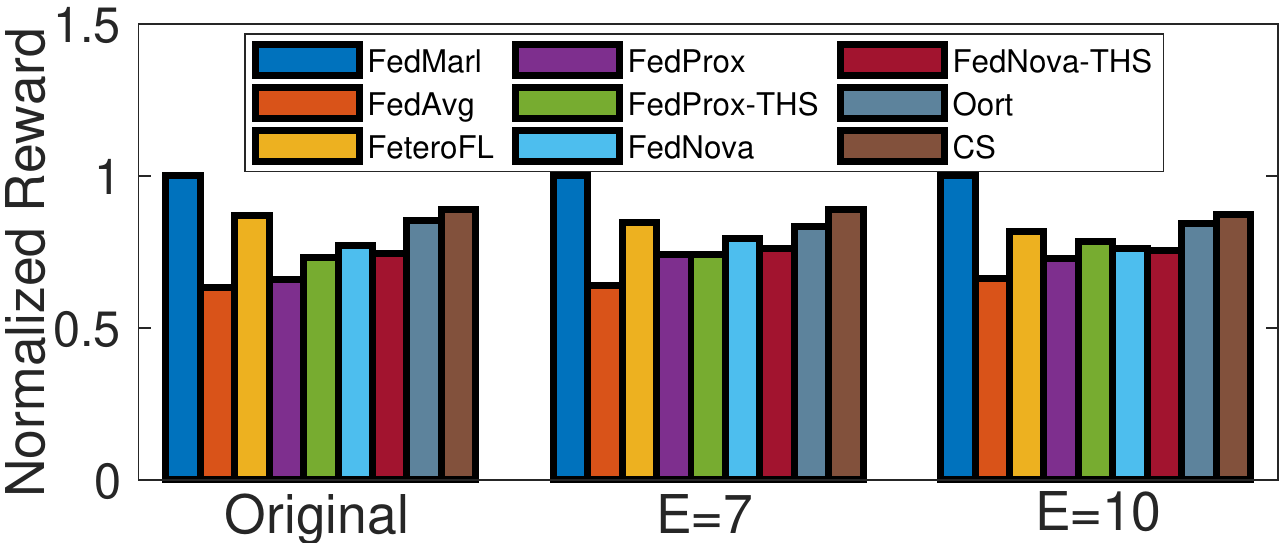}
    \end{subfigure}
    \begin{subfigure}{.33\textwidth}
        \centering
        \includegraphics[width=\linewidth]{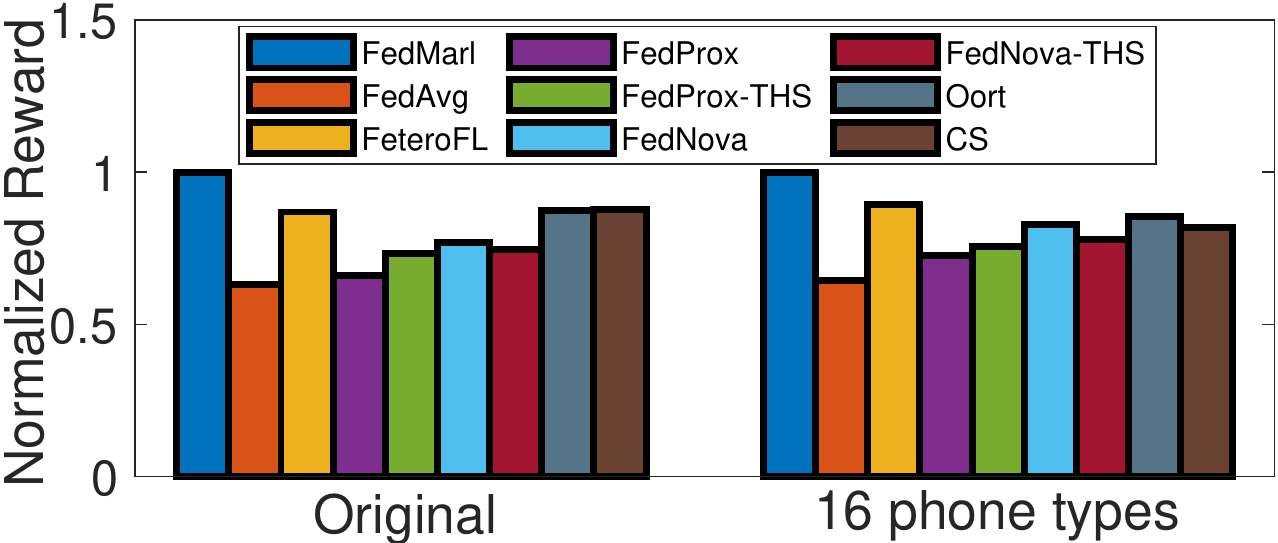}
    \end{subfigure}
    \caption{Performance of MARL agents under different system configurations. All the results are normalized with the performance of FedMarl.}
    \label{fig:transfer-performance}
    \end{minipage}
    \begin{minipage}{0.3\textwidth}
\resizebox{0.98\textwidth}{!}{
\begin{tabular}{c | c | c | c }
\hline
$[w_{1},w_{2},w_{3}]$ & LeNet & VGG6 & ResNet-18  \\ \hline
\multirow{ 3}{*}{$[1.0, 0.2, 0.1]$} & A: $98.15\%$ & A: $46.05\%$ & A: $98.43\%$\\
 & L: $1.0\times$ & L:$1.0\times$ & L:$1.0\times$ \\
 & B: $1.0\times$ & B:$1.0\times$ & B:$1.0\times$ \\ \hline
\multirow{ 3}{*}{$[1.0, 0.5, 0.3]$} & A: $97.88\%$ & A: $45.28\%$ & $98.03\%$\\
 &L: $0.82\times$ & L:$0.80\times$ & L:$0.80\times$\\
 &B: $0.83\times$ & B:$0.78\times$ & B:$0.86\times$\\ \hline
\end{tabular}}
\captionof{table}{'A','L','B' denote the accuracy, latency and communication cost. }
\label{table:ablation-study}
    \end{minipage}
\end{figure*}

\subsection{Generalizability of the MARL Agents}

\label{sec:performance-dynamic}
In practice, FL system configuration can vary over time. For example, client devices may join and quit the FL application over time, leading to a varying client pool size K. This may require separate MARL agents to be trained for each system configuration, leading to a significant MARL training cost. Next, we employ the MARL agents trained with the default settings described in the beginning of the evaluation section, and evaluate their performances under different conditions.

\subsubsection{Evaluation under Different System Conditions}
We first evaluate the FedMarl performance by varying the pool size $K$. Specifically, we utilize the MARL agents trained with $K=100$ and evaluate the performance under $K=400$ and $K=800$. Figure~\ref{fig:transfer-performance}(a) depicts the total reward (defined in equation~\ref{eqn:marl-reward}) of the algorithms on CIFAR-10. Remember that a higher reward indicates a better overall performance in terms of prediction accuracy, processing latency and bandwidth consumption. It can be seen that FedMarl outperforms the rest algorithms for each $K$. In particular, FedMarl achieves a $1.36\times$ and $1.32\times$ higher total reward than the other algorithms on average under $K=400$ and $K=800$, respectively. Similarly, Figure~\ref{fig:transfer-performance}(b) shows the performance of the algorithms under different amount of epochs $E$ for local training. FedMarl also obtains the best overall performance under different $E$. Finally, we modify the FL system configuration by introducing new types of client devices. Specifically, in additional to the eight devices shown in Table~\ref{table:client-option}, we introduce another six 
mobile devices to the device pool including iPhone 12, iPhone 7, Samsung Galaxy S21, Google Pixel 5, Samsung A11, Huawei P20 pro. We collect the latency traces for these new devices and evaluate the performances of the algorithms using the new device pool on CIFAR-10 (Figure~\ref{fig:transfer-performance}(c)). The results show that the superior performance of the MARL agents can translate across different system settings. This is because the MARL agents only take normalized version of the inputs (equation~\ref{eqn:state}) for generating the decisions, making them independent of a specific system configuration.

\subsubsection{Evaluation across Different DNN Architectures and Datasets}
Next, we evaluate the generalizability of MARL agents over different DNN architectures and datasets. In particular, we train the MARL agents with a source DNN architecture (e.g., VGG6 on CIFAR-10), and evaluate their performance using a target DNN (e.g., ResNet-18 on Fashion MNIST). Table~\ref{table:cross-dnn-evaluation} shows the normalized rewards of the MARL agents. We notice that the MARL agents achieve a superior performance across different DNNs and datasets in general. For example, the MARL agents trained on ResNet-18 with Fashion MNIST can obtain a reward of 0.965 on LeNet with MNIST, which is comparable with the performance of the MARL agents trained with LeNet from scratch (1.0). Additionally, by finetuning the MARL agents with 10 episodes on the target DNN (Numbers in the brackets in Table~\ref{table:cross-dnn-evaluation}), we notice further improvements on the reward. This demonstrates that the trained MARL agents can generalize to different DNN architectures and datasets.  

\subsection{Learned Strategy by the MARL Agents}
In this section, we take a closer look at the strategies learnt by the MARL agents. Figure~\ref{fig:scheduling} shows the decisions made by the MARL agents during the FL process for LSTM. In particular, we investigate how the client selection decisions are affected by the probing losses (Figure~\ref{fig:scheduling} (a)) and the probing training latencies (Figure~\ref{fig:scheduling} (b)) of the clients. We make the following observations. First, the MARL agents prefer to select the clients with both low probing loss and training latency. Second, the MARL agents occasionally select the clients with low probing loss for the better FL convergence, even though their latencies are high. For instance, the two blue dots pointed by the arrows in
Figure~\ref{fig:scheduling} represent a client device with low probing loss and high processing latency. The MARL agents apply the optimal criteria to achieve a trade off between the training accuracy and total processing latency objectives. Finally, relatively less number of clients are picked in the early stage of the FL training than the later stage. In Figure~\ref{fig:scheduling}, 3 and 7 (out of 10) clients are selected at the first and last round, respectively. One reason is that in the early stage of training, DNNs usually learn low-complexity (lower-frequency) functional components before learning more advanced features, with the former being more robust to noises and perturbations~\cite{xu2019frequency,rahaman2019spectral}. Therefore it is possible to use less data to train in the early stages, which further reduces processing latency and communication cost.

\subsection{Performance under Different Rewards}
In this section, we investigate the impact of relative importance $w_{1},w_{2},w_{3}$ in the reward function (equation~\ref{eqn:marl-reward}) on the performance of FedMarl. Specifically, besides the original setting with $[w_{1},w_{2},w_{3}] = [1.0,0.2,0.1]$,
we increase $w_{2}$ from 0.2 to 0.5 and $w_{3}$ from 0.1 to 0.3. This will force the MARL agents to learn a client selection algorithm with a lower processing latency and communication cost. As shown in Table~\ref{table:ablation-study}, we observe that increasing $w_{2}$ and $w_{3}$ will lead to a lower total processing latency and communication cost at the price of lower accuracy. This indicates that FedMarl can adjust its behavior based on the relative importance of the objectives, which enables the application designers to customize the FedMarl based on their preferences.

\section{Conclusion}
In this work, we present FedMarl, an MARL-based FL framework that makes intelligent client selection at run time. The evaluation results show that FedMarl outperforms the benchmark algorithms in terms of model accuracy, processing latency and communication cost under various FL conditions. 

\bibliography{aaai22}

\end{document}